\documentclass[man,floatsintext]{apa7}
\usepackage{array}
\usepackage{enumitem}
\usepackage{graphicx} 
\usepackage{longtable}
\usepackage{multirow}
\usepackage{multicol}
\usepackage{subcaption}
\usepackage{tikz}
\usepackage{adjustbox}
\usepackage{placeins}
\usepackage[style=apa, backend=biber]{biblatex}

\title{NCME2025 - Toward Subtrait-Level Model Explainability in Automated Writing Evaluation}
\date{March 2025}

\title{Toward Subtrait-Level Model Explainability in Automated Writing Evaluation}
\shorttitle{Toward Subtrait-Level Model Explainability}
\author{Alejandro Andrade-Lotero\textsuperscript{1}, Lee Becker\textsuperscript{2}, Joshua Southerland\textsuperscript{2}\textsuperscript{\textdagger}, Scott Hellman\textsuperscript{2}, Bradley Behan\textsuperscript{2}}
\affiliation{
    \textsuperscript{1} Research conducted while at Pearson Education\\
    \textsuperscript{2} {Pearson Education, Inc.}
}
\authornote{
    Correspondence concerning this article should be addressed to Lee Becker. \\
    Pearson affiliated authors can be reached at \texttt{<given>.<surname>@pearson.com}.  \\
    \textdagger Joshua Southerland can be reached at josh.southerland@pearson.com
}
\date{March 2025}

\begin{document}

\maketitle

\begin{abstract}
    \leavevmode Subtrait (latent-trait components) assessment presents a promising path toward enhancing transparency of automated writing scores.  We prototype explainability and subtrait scoring with generative language models and show modest correlation between human subtrait and trait scores, and between automated and human subtrait scores. Our approach provides details to demystify scores for educators and students. 
\end{abstract}

\section{Introduction}

We present a novel approach to enhancing the transparency of automated writing evaluation (AWE) systems. Leveraging the latest advances in generative language models (GLM), we developed a comprehensive evaluation of various facets of writing. Central to this system is a rubric, which breaks down latent traits into finer components called subtraits. For example, a common trait in the writing domain is Organization, which is composed of various subtraits (Introduction, Paragraph Strategies, Cohesion, and Conclusion). It is at this level that our system operates. In what follows, we examine the specific challenges associated with explaining AWE and trait-based assessment, and discuss our approach to granular, subtrait-driven analysis and its potential benefits.

\subsection{Explainability in AWE Systems}

Traditional AWE systems make use of natural language processing (NLP) techniques by extracting lexical, syntactical, and semantical properties of text \parencite{shermis_handbook_2013, huawei_systematic_2023, xue_towards_2024}. For instance, n-grams, part-of-speech tags, dependency parsing, lexical complexity, and coherence metrics, etc. The challenges of these traditional models are the posteriori finding of correlations between these features and writing scores \parencite{crossley_linguistic_2020}. While these NLP features serve as proxies for writing skills, they are not standalone explanations. These features do not directly explain the skills, and therefore require an additional layer of work to make them explainable. Recent advancements in Automatic Word Embedding (AWE) have moved towards transformer-based models \parencite{clark_electra_2020, devlin_bert_2019}, which represent documents within a joint syntactic and semantic space.  While useful for predicting scores, these black-box models do not give insight into specific performances of writing skills.  Our novel approach frames AWE as subtrait assessment to enhance explainability. 

\subsection{Trait-Based Writing Assessment}

Many rubrics represent a grouping of writing constructs \parencite{bacha_writing_2001, graham_formative_2015}. Depending on the stakeholder, the number of constructs can range as high as 5 or more, though these constructs tend to be expressed in different wordings by different stakeholders. This work decomposes high-level traits into fine-grained subskills referred to in this paper as ``subtraits''. For example, Transcend rubrics \parencite{edmetric_llc_evaluation_2020} assess a construct called \emph{Purpose and Organization} with four descriptors: \emph{Introductory Statement}, \emph{Organizational Structure of Paragraphs}, \emph{Cohesion and Transitions}, and \emph{Concluding Statement}.

\subsection{Writing Skills Tree}

We have developed a writing skills tree for middle school that decomposes trait-level skills into independent components or subtraits. An excerpt is shown in Table~\ref{tab:writing-skills-tree}. Each subtrait is accompanied by a short description and a rubric, and is tagged with relevant Common Core Standards. This granularity is designed to allow for flexible alignment to a variety of rubric traits while still ensuring the final rubric assesses the latent construct of writing proficiency \parencite{deane_cognitive_2008}.

\section{Research Questions}
\begin{enumerate}[labelindent=0pt, parsep=0pt, itemsep=0pt, labelsep=0.5em, topsep=0pt] 
    \item What are the advantages and disadvantages of modeling subtrait rather than trait scores?
    \item How reliably can subtraits be scored by human markers?
    \item What is the performance of zero-shot prompting for scoring subtraits with GLMs?
    \item Do extracted spans provide some degree of explainability to subtrait scores?
\end{enumerate}

\section{Methods}

This study consisted of three phases: 1) Developing the Writing Skills Tree (see Table~\ref{tab:writing-skills-tree}), 2) Scoring and annotating response data and 3) Automated subtrait scoring and evidence extraction using zero-shot prompting with GLMs.

\subsection{Data Sources and Collection}

Human markers were asked to score 225 constructed responses to informative/explanatory writing items using our subtraits rubric. These 225 responses were sampled across 45 different middle school prompts (5 responses per prompt). Unlike, traditional AWE scoring, responses and items were pooled into a single dataset to encourage generalization of scoring across a diverse set of prompts. 

In total 4 subject matter experts in English and Language Arts served as scorers to provide two reads per response to allow for analysis of inter-rater reliability (IRR). For each response the scorers were instructed to:
\begin{enumerate}[topsep=0pt, parsep=0pt, itemsep=0pt]
    \item Score responses for two (2) traits
    \item Score responses for eight (8) subtraits (4 subtrait per trait)
    \item Identify and highlight regions of the student response which support the subtrait score decision.
\end{enumerate}

The selected traits and subtraits were selected for the applicability to a wide variety of English and Language Arts assignments and for the prompt-agnostic nature.  

\subsection{Zero-shot Subtrait Scoring}

A goal of this work is to understand how capable GLMs are ``out of the box'' for automated subtrait scoring.  Recent advances in generative language modeling have shown an emergent property of reasoning beyond word prediction \parencite{kojima_large_2022} which allows GLMs to carry out tasks solely by way of instruction.  This zero-shot approach could greatly benefit AWE, as a GLM could produce both scores and explanations of those scores without requiring the training of a traditional supervised machine learning model.

\begin{figure}
    \caption{Example Prompt Configuration}
    \centering
    \includegraphics[width=0.9\linewidth]{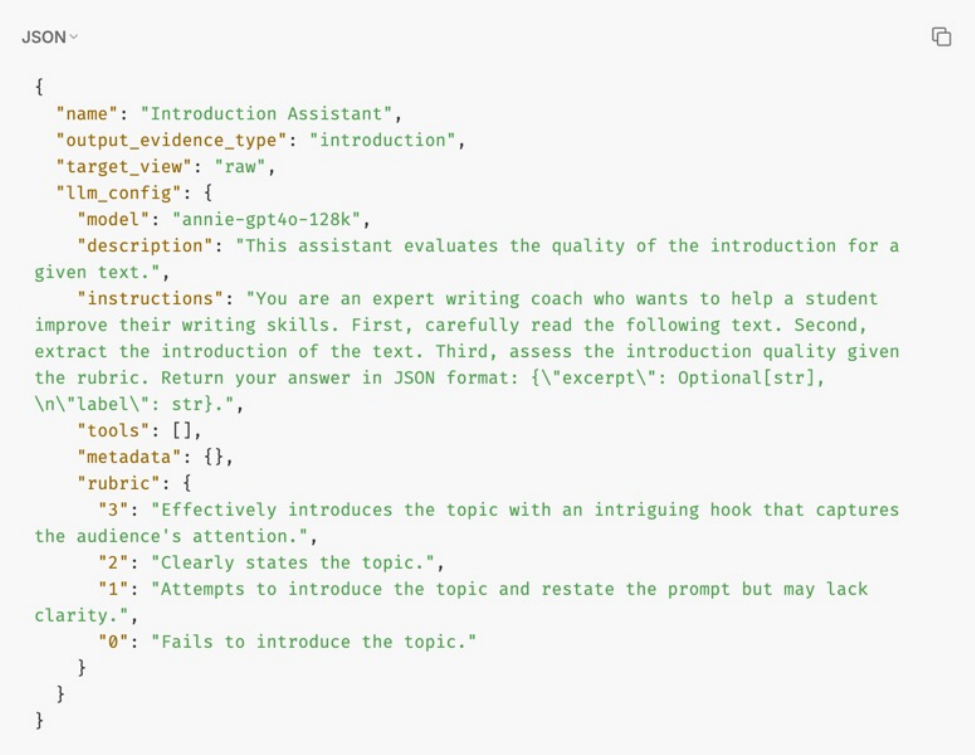}
    \label{fig:prompt-config}
\end{figure}

For this pilot study, subtrait scoring is framed as a zero-shot prompting technique that leverages item and rubric info to instruct and guide the GLM into providing scores.  An example configuration containing the GLM prompt is displayed in Figure~\ref{fig:prompt-config}. This example configures the GLM to extract and evaluate the use of introductory sentences.  OpenAI's GPT-4o mini deployed on an internal Microsoft Azure account was the sole GLM used for this work.  This decision was based on general performance of Open AI models along with practical concerns such as API access and data privacy policies.

The stochastic nature of inference in generative models introduces an variability into scoring that is not present with traditional supervised learning. 
For a given subtrait each response was processed 10 times with the relevant GLM context to produce 10 scores.  This redundant processing enables inquiry into the GLM models' consistency across runs.

\section{Results}

\subsection{Inter-Rater Reliability}

Inter-rater reliability metrics for traits and subtraits are listed in Table~\ref{tab:human-human-agreement}.  Quadratic weighted kappa (QWK) for the traits was mixed with lower agreement for \emph{Purpose and Organization} (0.338 QWK) and moderate agreement for \emph{Evidence and Elaboration} (0.537 QWK).

\begin{figure}[htbp]
  \caption{Purpose and Organization Subtraits Inter-Rater Confusion Matrices }
  \centering

  \begin{subfigure}[t]{0.45\textwidth}
    \centering
    \begin{tikzpicture}
      \node[anchor=south west, inner sep=0] at (0,0) {\includegraphics[width=\linewidth]{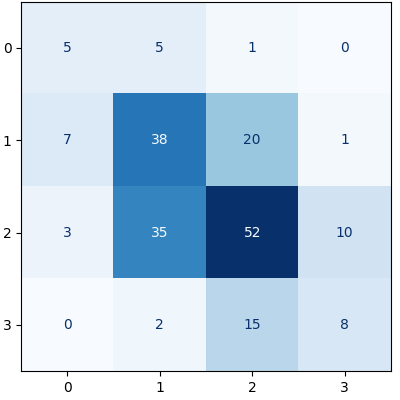}};
      \node[rotate=90] at (-0.5,2.1) {First Read};
      \node at (2.4,-0.3) {Second Read};
    \end{tikzpicture}
    \caption{Cohesion and Transitions}
  \end{subfigure}
  \hfill
  \begin{subfigure}[t]{0.45\textwidth}
    \centering
    \begin{tikzpicture}
      \node[anchor=south west, inner sep=0] at (0,0) {\includegraphics[width=\linewidth]{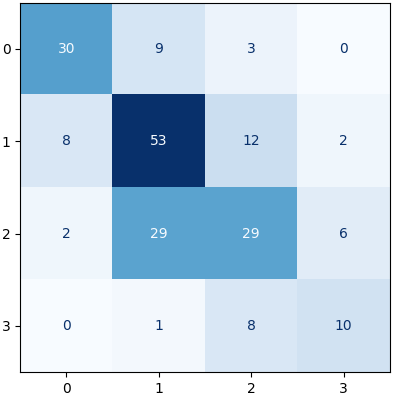}};
      \node[rotate=90] at (-0.5,2.1) {First Read};
      \node at (2.4,-0.3) {Second Read};
    \end{tikzpicture}
    \caption{Introduction of the Topic}
  \end{subfigure}

  \vspace{1em}

  \begin{subfigure}[t]{0.45\textwidth}
    \centering
    \begin{tikzpicture}
      \node[anchor=south west, inner sep=0] at (0,0) {\includegraphics[width=\linewidth]{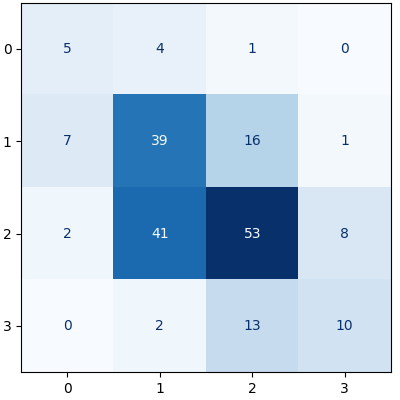}};
      \node[rotate=90] at (-0.5,2.1) {First Read};
      \node at (2.4,-0.3) {Second Read};
    \end{tikzpicture}
    \caption{Paragraph Organization Strategies}
  \end{subfigure}
  \hfill
  \begin{subfigure}[t]{0.45\textwidth}
    \centering
    \begin{tikzpicture}
      \node[anchor=south west, inner sep=0] at (0,0) {\includegraphics[width=\linewidth]{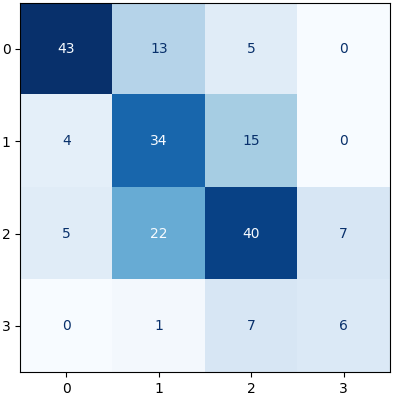}};
      \node[rotate=90] at (-0.5,2.1) {First Read};
      \node at (2.4,-0.3) {Second Read};
    \end{tikzpicture}
    \caption{Concluding Statement}
  \end{subfigure}
  \label{fig:human-human-confusion}

\end{figure}

Table ~\ref{tab:human-human-agreement} show moderate inter-rater agreement (0.4-0.6 QWK) for a majority of the subtrait scores.  For the \emph{Purpose and Organization} trait, IRR was lower than for subtrait scores, whereas for the \emph{Evidence and Elaboration} trait, IRR was roughly the same as for the subtraits.  Inspection of the confusion matrices (Figure ~\ref{fig:human-human-confusion}) shows that the majority of subtrait scores fell and confusion fell in the middle scorepoint ranges (SP1, SP2).  The one exception was for the \emph{Introduction of the Topic} and \emph{Concluding Statement} subtraits which had strong agreement at the bottom scorepoint (SP0).

Feedback from the human markers suggested that co-mingling responses from multiple item prompts made the task more challenging as it is different from operational scoring where scoring is done an item at a time.  Furthermore our scoring tool is unable to display the prompt context in line with the response.  We hypothesize this contributed to some of the discrepancies in IRR across subtraits.  For example, \emph{Introduction of the Topic} and \emph{Concluding Statement} are straightforward constructs to assess without knowing the prompt expectations, whereas \emph{Evidence and Elaboration} subtraits like \emph{Domain Specific Vocabulary} and \emph{Maintain a Formal Style} may require recall of the prompt text and passage readings to know what is appropriate.  For future data collection, inter-rater reliability could be improved by allowing raters to work through a single item's responses at a time.

Trait scores and subtrait scores are highly correlated for both of the evaluated traits (Table~\ref{tab:trait-subtrait-correlation}).  Pearson-r Correlation between trait and subtrait scores was computed averaging the first and second reads for each trait and subtrait, and then averaging the four average subtrait scores.

\subsection{Scoring Performance}

\begin{figure}[htbp]
  \caption{Purpose and Organization Subtraits Model Confusion Matrix - True Score is the rounded average of the markers' first and second reads.  Predicted Score is the GLM model prediction.  Individual cells represent the average over multiple runs of the GLM}
  \centering

  \begin{subfigure}[t]{0.45\textwidth}
    \centering
    \begin{tikzpicture}
      \node[anchor=south west, inner sep=0] at (0,0) {\includegraphics[width=\linewidth]{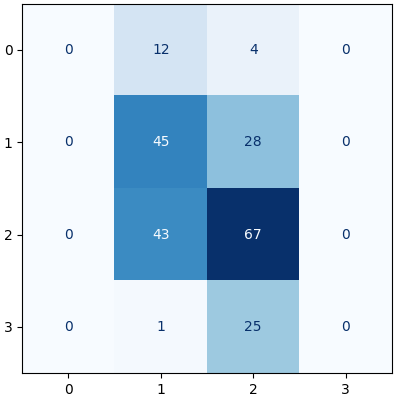}};
      \node[rotate=90] at (-0.5,2.1) {True Score};
      \node at (2.4,-0.3) {Predicted Score};
    \end{tikzpicture}
    \caption{Cohesion and Transitions}
  \end{subfigure}
  \hfill
  \begin{subfigure}[t]{0.45\textwidth}
    \centering
    \begin{tikzpicture}
      \node[anchor=south west, inner sep=0] at (0,0) {\includegraphics[width=\linewidth]{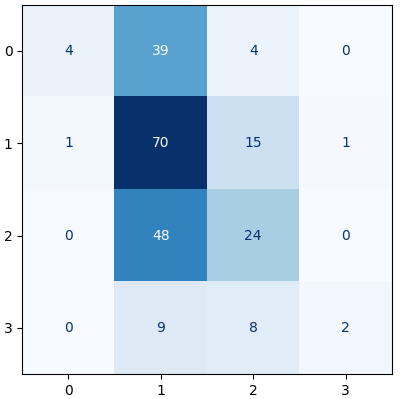}};
      \node[rotate=90] at (-0.5,2.1) {True Score};
      \node at (2.4,-0.3) {Predicted Score};
    \end{tikzpicture}
    \caption{Introduction of the Topic}
  \end{subfigure}

  \vspace{1em}

  \begin{subfigure}[t]{0.45\textwidth}
    \centering
    \begin{tikzpicture}
      \node[anchor=south west, inner sep=0] at (0,0) {\includegraphics[width=\linewidth]{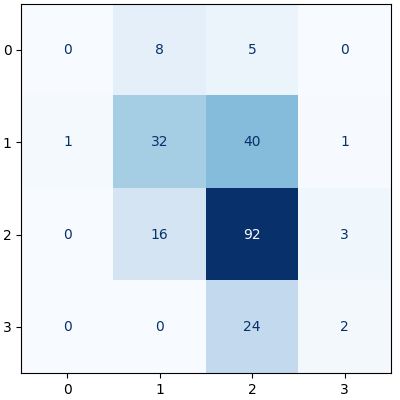}};
      \node[rotate=90] at (-0.5,2.1) {True Score};
      \node at (2.4,-0.3) {Predicted Score};
    \end{tikzpicture}
    \caption{Paragraph Organization Strategies}
  \end{subfigure}
  \hfill
  \begin{subfigure}[t]{0.45\textwidth}
    \centering
    \begin{tikzpicture}
      \node[anchor=south west, inner sep=0] at (0,0) {\includegraphics[width=\linewidth]{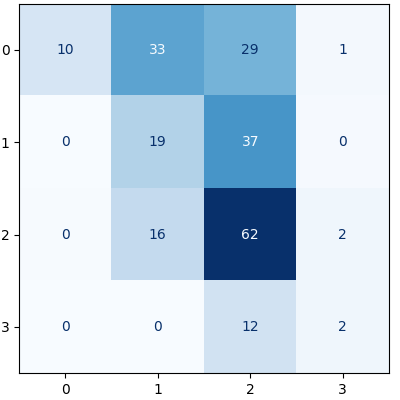}};
      \node[rotate=90] at (-0.5,2.1) {True Score};
      \node at (2.4,-0.3) {Predicted Score};
    \end{tikzpicture}
    \caption{Concluding Statement}
  \end{subfigure}

  \label{tab:human_model_org_confusion}
\end{figure}
\begin{figure}[htbp]
  \caption{Evidence and Elaboration Subtraits Model Confusion Matrix - True Score is the first readers' score.  Predicted Score is the LLM model prediction.  Individual cells represent the average over multiple runs of the LLM}
  \centering

  \begin{subfigure}[t]{0.45\textwidth}
    \centering
    \begin{tikzpicture}
      \node[anchor=south west, inner sep=0] at (0,0) {\includegraphics[width=\linewidth]{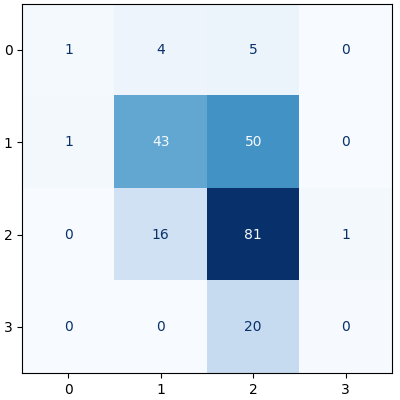}};
      \node[rotate=90] at (-0.5,2.1) {True Score};
      \node at (2.4,-0.3) {Predicted Score};
    \end{tikzpicture}
    \caption{Domain Specific Vocabulary}
  \end{subfigure}
  \hfill
  \begin{subfigure}[t]{0.45\textwidth}
    \centering
    \begin{tikzpicture}
      \node[anchor=south west, inner sep=0] at (0,0) {\includegraphics[width=\linewidth]{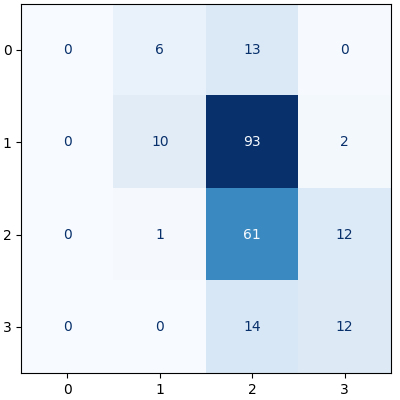}};
      \node[rotate=90] at (-0.5,2.1) {True Score};
      \node at (2.4,-0.3) {Predicted Score};
    \end{tikzpicture}
    \caption{Explanation of Main Points}
  \end{subfigure}

  \vspace{1em}

  \begin{subfigure}[t]{0.45\textwidth}
    \centering
    \begin{tikzpicture}
      \node[anchor=south west, inner sep=0] at (0,0) {\includegraphics[width=\linewidth]{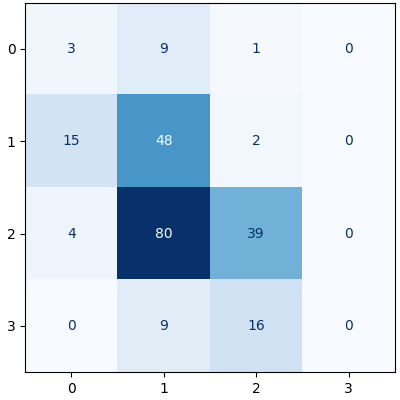}};
      \node[rotate=90] at (-0.5,2.1) {True Score};
      \node at (2.4,-0.3) {Predicted Score};
    \end{tikzpicture}
    \caption{Facts and Quotations}
  \end{subfigure}
  \hfill
  \begin{subfigure}[t]{0.45\textwidth}
    \centering
    \begin{tikzpicture}
      \node[anchor=south west, inner sep=0] at (0,0) {\includegraphics[width=\linewidth]{img/evidence/avg_model-fr_human-maintain_a_formal_style_cropped.png}};
      \node[rotate=90] at (-0.5,2.1) {True Score};
      \node at (2.4,-0.3) {Predicted Score};
    \end{tikzpicture}
    \caption{Maintain a Formal Style}
  \end{subfigure}
  \label{tab:human_model_evidence_confusion}

\end{figure}

Because the data collection did not include adjudication, the first read scores are considered the $y\_true$ for evaluation of GLM output.  QWK and Exact Agreement (Table~\ref{tab:human-model-agreement} were computed for each subtrait and included all predictions from the 10 runs of the GLM.  Across all subtraits the models exhibited fair agreement (QWK (0.2, 0.4]).  While the model agreements fall well below the typical 0.70 QWK required for high stakes assessment, several of the \emph{Evidence and Elaboration} subtraits had agreement similar to the human-human raters. Closer inspection of the confusion matrices (Figures~\ref{tab:human_model_org_confusion}, ~\ref{tab:human_model_evidence_confusion}) and classification metrics (Tables~\ref{tab:organization-classification_reports}, \ref{tab:evidence-classification_reports}) highlight the GLM's poor Recall for responses in the lowest and highest scorepoint rage. Seven out of ten of the subtraits had scorepoints with recall of 0.10 or less.  

The mean and standard deviation of Mean Absolute Error (MAE), Root Mean Square Error (RMSE) along with Krippendorff's Alpha \parencite{krippendorff2009testing} were computed to quantify the GLM subtrait scores consistency between successive runs on the same response.  Mean MAE values indicate that for a given response the GLM subtrait score prediction will vary by by less than 0.3 of a scorepoint and between responses that deviation is less than 0.025.  Similarly, the RMSE mean and std values give a hint at the amount that QWK will vary between random runs.  Computation of Krippendorff Alpha utilized the ordinal parameters and treated each model subtrait score as the output of an individual rater.  The high Krippendorff Alpha values confirm that the the GLM model is consistent in rating.

\subsection{Extracted Subtrait Evidence}

In addition to requesting subtrait scores the GLM prompt configuration included instructions to extract subtrait relevant regions (a.k.a. span evidence) from the student response.  These spans provide an additional layer of explainability that can help contextualize what text contributed to the subtrait scoring decision. Tables~\ref{tab:example_org_evidence} and \ref{tab:example_evi_evidence} present example span evidence selected by both human markers and the subtrait models. The evidence in these tables is derived from responses where there was agreement between human markers and the model on the subtrait scores to better contrast differences in behaviors.

The example \emph{Introduction of the Topic} evidence contains a high degree of overlap between human and model selections, though the human marker opted to select more of the first paragraph.  The GLM instead selected text that came after the introduction.  Overproduction of evidence was a recurring theme across many subtraits.  In \emph{Cohesion and Transitions}, the human selected individual transition words, whereas the model selected entire sentences containing transition words.

The \emph{Paragraph Organization Strategies} subtrait does not naturally lend itself to interpretable region selection.  In the example in Table~\ref{tab:example_org_evidence} the human marker selected the entire first paragraph, whereas the GLM model create span evidence for each paragraph in the response.  This is likely because the latent skill for this subtrait relates more to how the paragraph is structured than any specific content. 

With \emph{Concluding Statement} the behaviors greatly differ between high and low scoring responses.  Responses scoring 0, typically lack a conclusion, so no evidence was selected by the human marker, whereas the GLM model selected all the paragraph text for the response.  Conversely with the high score both human and model selected the entire conclusion.

The GLM model selected multiple, single words for the \emph{Domain Specific Vocabulary}.  Like in the example, the human markers often highlighted a single sentence that contains vocabulary.  With \emph{Explanation of Main Points} human and model extracted very different span evidence, but both seem plausible or justified. The \emph{Facts and Quotations} shows a common theme of having overlap, but also producing additional, less useful spans. 


\section{Implications for Future Research}


    What are the advantages and disadvantages of modeling subtrait rather than trait scores?
    How reliably can subtraits be scored by human markers?
    What is the performance of zero-shot prompting for scoring subtraits with GLMs?
    Do extracted spans provide some degree of explainability to subtrait scores?

The results of this pilot study found that human markers exhibit moderate inter-rater reliability in scoring subtraits.  Potential ways to improve human scoring reliability include further refinement of subtrait rubrics, additional marker training, and structuring the task and ordering of responses to minimize context switching.  These finding also suggest that successful subtrait scoring will depend on framing the skill assessment in a way that can work in a more prompt-independent manner as the usual mechanisms for calibrating to a single item do not scale.

Subtrait scores produced by zero-shot prompting of GLM models exhibited fair performance and for some subtraits agreement approached human-human agreement. For the subtrait scoring tasks, the GLM model output showed consistency between multiple passes over the same response.  If this prompting approach were to be used in a real application, fixing the model's initialization seed could further reduce variability. 

While the zero-shot approach is not yet viable for high stakes settings, the potential for large language models lies more in their ability to be used flexibly to assess a variety of skills and to augment existing trait scores with additional, targeted analyses about the student's writing.  In a lower stakes setting, these subtrait scores can inform feedback strategies and serve as features to personalized learning and recommender engines.

For future work we plan to investigate how subtrait scores and associated evidence can be utilized
to train an explanatory model for trait scores.  We also would like to collect more data and carry out additional experiments to explore the effect of prompt engineering, few-shot learning and fine-tuning on subtrait score accuracy.

Overall, the GLM-based subtrait scoring and evidence extraction presents an encouraging direction for enhancing automated writing evaluation with additional, targeted and human-interpretable information.

\printbibliography

\pagebreak

\section{Figures and Tables}

\begin{table}[ht]
    \caption{Trait and Subtrait Inter-Rater Reliability}
    \centering
    \begin{tabular}{ll|c|c}
        Trait  & Subtrait & QWK & Exact\\
        \hline
        Purpose and Organization & -          & 0.338 & 0.378\\
        \hline
        & Cohesion and Transitions            & 0.494 & 0.510\\
        & Introduction of the Topic           & 0.673 & 0.604 \\
        & Paragraph Organization Strategies   & 0.518 & 0.530 \\
        & Concluding Statement                & 0.675 & 0.609 \\
        \hline
        Evidence and Elaboration & -    & 0.537  & 0.478 \\
        \hline
        & Domain Specific Vocabulary    & 0.384 & 0.455\\
        & Explanation of Main Points    & 0.559 & 0.534\\
        & Facts and Quotations          & 0.582 & 0.550\\
        & Maintain a Formal Style       & 0.464 & 0.460
    \end{tabular}
    \label{tab:human-human-agreement}
\end{table}

\begin{table}[h]
    \caption{Trait-Subtrait Correlation}
    \label{tab:trait-subtrait-correlation}
    \centering
    \begin{tabular}{l|c}
         Trait & \emph{r} \\
         \hline
         Purpose and Organization & 0.718 \\
         Evidence and Elaboration & 0.767 \\
         \hline
    \end{tabular}
\end{table}

\begin{figure}[htbp]
  \caption{Evidence and Elaboration Subtraits Inter-Rater Confusion Matrices }
  \centering

  \begin{subfigure}[t]{0.45\textwidth}
    \centering
    \begin{tikzpicture}
      \node[anchor=south west, inner sep=0] at (0,0) {\includegraphics[width=\linewidth]{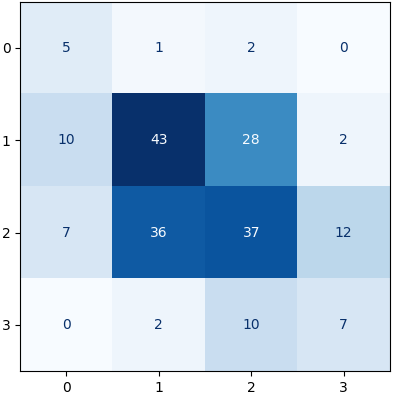}};
      \node[rotate=90] at (-0.5,2.1) {First Read};
      \node at (2.4,-0.3) {Second Read};
    \end{tikzpicture}
    \caption{Domain Specific Vocabulary}
  \end{subfigure}
  \hfill
  \begin{subfigure}[t]{0.45\textwidth}
    \centering
    \begin{tikzpicture}
      \node[anchor=south west, inner sep=0] at (0,0) {\includegraphics[width=\linewidth]{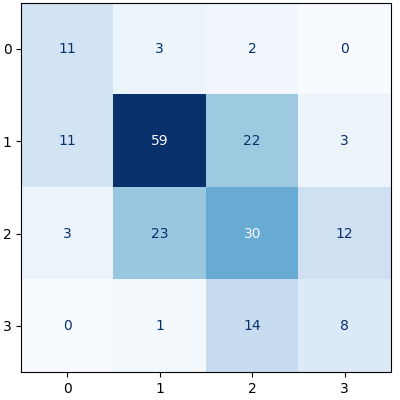}};
      \node[rotate=90] at (-0.5,2.1) {First Read};
      \node at (2.4,-0.3) {Second Read};
    \end{tikzpicture}
    \caption{Explanation of Main Points}
  \end{subfigure}

  \vspace{1em}

  \begin{subfigure}[t]{0.45\textwidth}
    \centering
    \begin{tikzpicture}
      \node[anchor=south west, inner sep=0] at (0,0) {\includegraphics[width=\linewidth]{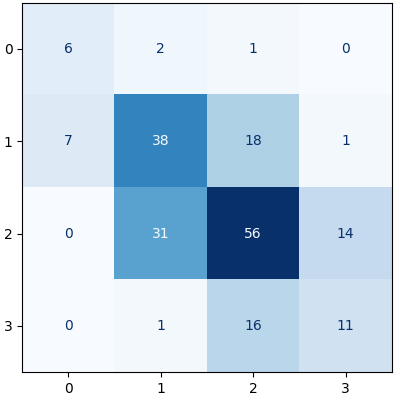}};
      \node[rotate=90] at (-0.5,2.1) {First Read};
      \node at (2.4,-0.3) {Second Read};
    \end{tikzpicture}
    \caption{Facts and Quotations}
  \end{subfigure}
  \hfill
  \begin{subfigure}[t]{0.45\textwidth}
    \centering
    \begin{tikzpicture}
      \node[anchor=south west, inner sep=0] at (0,0) {\includegraphics[width=\linewidth]{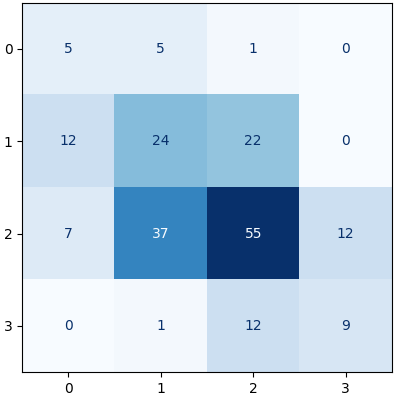}};
      \node[rotate=90] at (-0.5,2.1) {First Read};
      \node at (2.4,-0.3) {Second Read};
    \end{tikzpicture}
    \caption{Maintain a Formal Style}
  \end{subfigure}

\end{figure}

\begin{table}[ht]
    \caption{Human model subtrait scoring agreement}
    \centering
    \begin{tabular}{ll|c|c}
        Trait  & Subtrait & QWK & Exact \\
        \hline
        Purpose and Organization 
        & Cohesion and Transitions            & 0.333 & 0.496\\
        & Introduction of the Topic           & 0.291 & 0.445\\
        & Paragraph Organization Strategies   & 0.368 & 0.561\\
        & Concluding Statement                & 0.325 & 0.413\\
        \hline
        Evidence and Elaboration 
        & Domain Specific Vocabulary    & 0.342 & 0.559\\
        & Explanation of Main Points    & 0.282 & 0.368\\
        & Facts and Quotations          & 0.489 & 0.553\\
        & Maintain a Formal Style       & 0.333 & 0.399
    \end{tabular}
    \label{tab:human-model-agreement}
\end{table}

\begin{table}[ht]
\caption{Classification Reports for Purpose and Organization Subtrait Model Predictions}
\label{tab:organization-classification_reports}
\centering
\begin{tabular}{|l|l|c|c|c|}
\hline
\textbf{Subtrait} & \textbf{SP} & \textbf{Precision} & \textbf{Recall} & \textbf{F1-score} \\
\hline
\textbf{Cohesion and Transitions} 
& 0 & 0.00 & 0.00 & 0.00 \\
& 1 & 0.45 & 0.62 & 0.52 \\
& 2 & 0.54 & 0.61 & 0.57 \\
& 3 & 1.00 & 0.00 & 0.00 \\
\cline{2-5}
& avg & 0.50 & 0.31 & 0.27  \\
\hline
\textbf{Introduction of the Topic} 
& 0 & 0.75 & 0.09 & 0.17 \\
& 1 & 0.42 & 0.80 & 0.55 \\
& 2 & 0.47 & 0.33 & 0.39 \\
& 3 & 0.71 & 0.11 & 0.18 \\
\cline{2-5}
& avg & 0.59 & 0.33 & 0.32 \\
\hline
\textbf{Paragraph Organization Strategies} 
& 0 & 0.00 & 0.00 & 0.00 \\
& 1 & 0.57 & 0.44 & 0.49 \\
& 2 & 0.57 & 0.83 & 0.68 \\
& 3 & 0.34 & 0.08 & 0.13 \\
\cline{2-5}
& avg & 0.37 & 0.34 & 0.32 \\
\hline
\textbf{Concluding Statement} 
& 0 & 0.94 & 0.14 & 0.24 \\
& 1 & 0.28 & 0.33 & 0.30 \\
& 2 & 0.44 & 0.77 & 0.56 \\
& 3 & 0.37 & 0.16 & 0.22 \\
\cline{2-5}
& avg & 0.51 & 0.35 & 0.33  \\
\hline
\end{tabular}
\end{table}

\begin{table}[ht]
\caption{Classification Reports for Evidence and Elaboration Subtrait Model Predictions}
\label{tab:evidence-classification_reports}
\centering
\begin{tabular}{|l|l|c|c|c|}
\hline
\textbf{Subtrait} & \textbf{SP} & \textbf{Precision} & \textbf{Recall} & \textbf{F1-score} \\
\hline
\textbf{Domain Specific Vocabulary}
& 0 & 0.40 & 0.10 & 0.16 \\
& 1 & 0.68 & 0.45 & 0.54 \\
& 2 & 0.52 & 0.82 & 0.64 \\
& 3 & 0.24 & 0.02 & 0.04 \\
\cline{2-5}
& avg & 0.46 & 0.35 & 0.35 \\
\hline
\textbf{Explanation of Main Points}
& 0 & 1.00 & 0.01 & 0.03 \\
& 1 & 0.59 & 0.10 & 0.17 \\
& 2 & 0.34 & 0.82 & 0.48 \\
& 3 & 0.44 & 0.45 & 0.44 \\
\cline{2-5}
& avg & 0.59 & 0.35 & 0.28 \\
\hline
\textbf{Facts and Quotations} 
& 0 & 1.00 & 0.08 & 0.15 \\
& 1 & 0.61 & 0.34 & 0.44 \\
& 2 & 0.56 & 0.77 & 0.65 \\
& 3 & 0.46 & 0.46 & 0.46 \\
\cline{2-5}
& avg & 0.66 & 0.41 & 0.42 \\
\hline
\textbf{Maintain a Formal Style} 
& 0 & 0.14 & 0.23 & 0.17 \\
& 1 & 0.33 & 0.74 & 0.46 \\
& 2 & 0.67 & 0.32 & 0.43 \\
& 3 & 1.00 & 0.00 & 0.00 \\
\cline{2-5}
& avg & 0.54 & 0.32 & 0.27  \\
\hline
\end{tabular}
\end{table}

\begin{table}[ht]
    \caption{Model batch scoring deviation. MAE is the mean and std dev of Mean Absolute Error (Deviation) across pairs of LLM runs.  RMSE is the mean and std dev of the Root Mean Square Error (Deviation) across pairs of LLM runs.  Krippendorff's Alpha ($\alpha$) is computed across all batches.}
    \footnotesize
    \begin{tabular}{ll|l|l|c}
        \textbf{Trait}  & \textbf{Subtrait} & \textbf{MAE} & \textbf{RMSE} & \textbf{$\alpha$} \\
        \hline
        \multirow{2}{2.5cm}{Purpose and Organization}
        & Cohesion And Transitions            & $0.135\pm 0.018$ & $0.367\pm 0.024$ & 0.729 \\
        & Introduction of the Topic           & $0.175\pm 0.019$ & $0.417\pm 0.024$ & 0.630 \\
        & Paragraph Organization Strategies   & $0.150\pm 0.018$ & $0.387\pm 0.024$ & 0.684 \\
        & Concluding Statement                & $0.275\pm 0.025$ & $0.547\pm 0.034$ & 0.604 \\
        \hline
        \multirow{2}{2.5cm}{Evidence and Elaboration}
        & Domain Specific Vocabulary          & $0.197\pm 0.020$ & $0.444\pm 0.024$ & 0.592 \\
        & Explanation of Main Points          & $0.134\pm 0.020$ & $0.364\pm 0.027$ & 0.664 \\
        & Facts and Quotations                & $0.173\pm 0.021$ & $0.415\pm 0.026$ & 0.738 \\
        & Maintain a Formal Style             & $0.162\pm 0.021$ & $0.401\pm 0.026$ & 0.753 \\
    \end{tabular}
    \label{tab:batch-batch-agreement}
\end{table}

\begin{table}[]
    \caption{Extracted evidence from \emph{Organization and Purpose} subtraits}
    \scriptsize
    \centering
    \begin{tabular}{p{2.3cm}|c|p{5.5cm}|p{5.5cm}}
        Subtrait & Score & Human Evidence & Model Evidence \\ 
        \hline
        Introduction of the Topic
        & 3 
        & 1. \textit{When someone says bubbling hot mud springs, geysers shooting water into the air, herds of bison. Did you think of Yellowstone National Park? If you did you are right. Yellowstone is the nation's oldest and most familiar national park. \ldots and preserving the park.} 
        & 1. \textit{When someone says bubbling hot mud springs, geysers shooting water into the air, herds of bison. Did you think of Yellowstone National Park? If you did you are right. Yellowstone is the nation's oldest and most familiar national park.} \newline
          2. \textit{But remember there are some creatures that \ldots} \newline \\
        \hline
        Cohesion and Transitions
        & 2
        & 1. \textit{Gradually}\newline
          2. \textit{Unlike}\newline
          3. \textit{The article lastly said} \newline
          4. \textit{Also} 
        & 1. \textit{The benefits and disadvantages of REDACTED in the early 1900s.} \newline
          2. \textit{In the text \ldots And when REDACTED created the \ldots} \newline
          \ldots \newline
          6. \textit{The article lastly said \ldots} \newline
          7. \textit{Both texts have great beefits and disadvantages \ldots} \\
        \hline
        Paragraph Organization Strategies
        & 1
        & 1. (PARAGRAPH 1)
        & 1. (PARAGRAPH 1) \newline
          2. (PARAGRAPH 2) \newline
          \ldots \newline
          6. (PARAGRAPH 6) \\
        \hline
        Concluding Statement
        & 0
        & 1. (NO EVIDENCE SELECTED)
        & 1. (PARAGRAPH 1) \newline
          2. (PARAGRAPH 2) \newline
          3. (PARAGRAPH 3) \\
        \hline
        Concluding Statement
        & 3
        & 1. \textit{All in all, REDACTED and REDACTED both have the same message of with everything you want, it is imposible to be happy in life but with smaller things, it makes it much easier. \ldots the author demonstrates what happens if you do.}
        & 1. \textit{Both of these stories have the same idea of \ldots All in all, REDACTED and REDACTED both have the same message of with everything you want, it is imposible to be happy in life but with smaller things, it makes it much easier. \ldots the author demonstrates what happens if you do.} \\
        
    \end{tabular}
    \label{tab:example_org_evidence}
\end{table}

\begin{table}[]
    \caption{Extracted evidence from \emph{Evidence and Elaboration} subtraits}
    \scriptsize
    \centering
    \begin{tabular}{p{2.3cm}|c|p{5.5cm}|p{5.5cm}}
        Subtrait & Score & Human Evidence & Model Evidence \\ 
        \hline
        Domain Specific Vocabulary 
        & 1 
        & 1. \textit{talking about nature and a national park}
        & 1. \textit{litter} \newline
          2. \textit{wild life} \newline
          \ldots \newline
          3. \textit{water fal} \newline \\
      \hline
        Explanation of Main Points
        & 3
        & 1. \textit{As you can see even the tribes wanted \ldots  But with the emojis they can better understand what each word means so they can truly understand the language and culture.}
        & 1. \textit{There are many ways to keep these languages alive. We could teach them in class, ask the government for help, and even use emojis.} \newline
          2. \textit{Teaching about these languages and cultures in the class room is very important. \ldots} \newline
          3. \textit{One more idea on how people could keep cultures and languages alive is by \ldots}
        \\
        \hline
        Facts and Quotations
        & 2
        & 1. \textit{If you look apon A small grassy hill in the northern part of \ldots pays tribute to his gift.}
        & 1. \textit{The narrator is telling us how A man named \ldots by turning him into ash.} \newline
          2. \textit{When the day of preformance came \ldots didn't stop him from singing.} \newline
          \ldots \newline
          4. \textit{If you look apon A small grassy hill in the northern part of \ldots pays tribute to his gift. - Narrator} \newline \\
        \hline
        Maintain a Formal Style
        & 1 
        & 1. \textit{to conclude all of these studies ans speculations and facts show one thing.}
        & 1. \textit{Teenagers can not multitask effectively. \ldots are still really really bad.} \newline
          2. \textit{in the article ``REDACTED'' \ldots is a huge Reasoned judgment.} \newline
          \ldots \newline
          8. \textit{And multitasking tends to cause mistakes and flaws in both tasks.} \\
        
    \end{tabular}
    \label{tab:example_evi_evidence}
\end{table}

\FloatBarrier  
{
\scriptsize
\begin{longtable}{p{2cm}p{2cm}p{2.5cm}p{3.5cm}p{2.5cm}}

    \caption{Example extract of Pearson Writing Skills Tree for Middle School.}\\
    
        \textbf{Skill} & \textbf{Subtrait} & \textbf{Description} & \textbf{Rubric} & \textbf{CCCS Tags} \\
        \hline
        Purpose and Organization 
        & Introduction of the Topic 
        & The student effectively introduces the topic or thesis and catches the readers' interest.
        &\begin{enumerate}[start=0, left=0pt, labelindent=0pt, itemindent=0pt, itemsep=0pt, labelsep=0.5em, topsep=0pt] 
            \item Fails to introduce the topic. 
            \item Attempts to introduce the topic and restate the prompt but may lack clarity. 
            \item Clearly states the topic or thesis. 
            \item Effectively introduces the topic with an intriguing hook that captures the audience's attention. 
        \end{enumerate}
        &\begin{itemize}[left=0pt, itemsep=0pt, topsep=0pt, partopsep=0pt, parsep=0pt] 
            \item CCSS.ELA-Literacy.W.6.2.a
            \item CCSS.ELA-Literacy.W.7.2.a
            \item CCSS.ELA-Literacy.W.8.2.a
        \end{itemize}\\
        
        & Organization Strategies
        & Paragraphs are organized logically and coherently.
        &\begin{enumerate}[start=0, left=0pt, labelindent=0pt, topsep=0pt, partopsep=0pt, parsep=0pt, itemindent=0pt, itemsep=0pt, labelsep=0.5em, topsep=0pt] 
            \item Lacks organizational structure.
            \item The text shows some grouping of ideas but lacks a clear organizational structure.
            \item Inconsistently applies organizational strategies in paragraphs such as definition, classification, comparison/contrast, and cause/effect.
            \item Effectively utilizes strategies like definition, classification, comparison/contrast, and cause/effect to organize paragraphs.
        \end{enumerate}
        &\begin{itemize}[left=0pt, itemsep=0pt, topsep=0pt, partopsep=0pt, parsep=0pt] 
            \item CCSS.ELA-Literacy.W.6.2.a
            \item CCSS.ELA-Literacy.W.7.2.a
            \item CCSS.ELA-Literacy.W.8.2.a
        \end{itemize}\\

        & Cohesion and Transitions
        & Logical and organized thematic progression of ideas
        &\begin{enumerate}[start=0, left=0pt, labelindent=0pt, itemindent=0pt, itemsep=0pt, labelsep=0.5em, topsep=0pt] 
            \item Fails to connect sentences and ideas. 
            \item Uses little or no transitional strategies between sentences, ideas, and paragraphs. 
            \item Uses some and/or repetitive transitional words, phrases, and/or clauses to connect sentences, ideas, and paragraphs. 
            \item Effectively uses a variety of connective words, phrases, and/or clauses to transition between sentences, ideas, and paragraphs.
        \end{enumerate}
        &\begin{itemize}[left=0pt, itemsep=0pt, topsep=0pt, partopsep=0pt, parsep=0pt] 
            \item CCSS.ELA-Literacy.W.6.1.c
            \item CCSS.ELA-Literacy.W.7.1.c
            \item CCSS.ELA-Literacy.W.8.1.c
            \item CCSS.ELA-Literacy.W.6.2.c
            \item CCSS.ELA-Literacy.W.7.2.c
            \item CCSS.ELA-Literacy.W.8.2.c
        \end{itemize}\\

        & Concluding Statement
        & Provide a concluding statement or section that follows from and supports the argument presented.
        &\begin{enumerate}[start=0, left=0pt, labelindent=0pt, itemindent=0pt, itemsep=0pt, labelsep=0.5em, topsep=0pt] 
            \item Lacks a statement that concludes and summarizes the main points.
            \item Includes a statement that represents an ending, but may not accurately summarize the main points.
            \item Communicates a conclusion but may lack a sense of completeness.
            \item Effectively concludes in a logical and unified manner, and gives a sense of completeness beyond restating the introductory statement.
        \end{enumerate}
        &\begin{itemize}[left=0pt, itemsep=0pt, topsep=0pt, partopsep=0pt, parsep=0pt] 
            \item CCSS.ELA-Literacy.W.6.2.f
            \item CCSS.ELA-Literacy.W.7.2.f
            \item CCSS.ELA-Literacy.W.8.2.f
            \item CCSS.ELA-Literacy.W.6.1.e
            \item CCSS.ELA-Literacy.W.7.1.e
            \item CCSS.ELA-Literacy.W.8.1.e
            \item CCSS.ELA-Literacy.W.6.3.e
            \item CCSS.ELA-Literacy.W.7.3.e
            \item CCSS.ELA-Literacy.W.8.3.e
        \end{itemize}\\
    \label{tab:writing-skills-tree}
\end{longtable}
}

\end{document}